\documentclass[letterpaper,10pt,conference]{ieeeconf}
\usepackage{graphicx}
\usepackage{url}
\usepackage{amsmath,amssymb}
\usepackage{array}
\usepackage{multirow}
\usepackage[table]{xcolor}
\definecolor{rowblue}{RGB}{230,245,250}
\definecolor{linkblue}{HTML}{0050A0}
\usepackage{placeins}
\usepackage{afterpage}
\makeatletter
\let\NAT@parse\undefined
\makeatother
\usepackage[colorlinks=true, citecolor=linkblue, linkcolor=linkblue, urlcolor=linkblue]{hyperref}
\IEEEoverridecommandlockouts
\title{\LARGE \bf GraphPoint: Semantic Entity Graphs and Point Trajectories for Compositional Robot Manipulation}

\author{
Kang Luo, Hesheng Wang$^\dagger$
\thanks{Kang Luo and Hesheng Wang are with IRMV Lab, the Department of Automation, Shanghai Jiao Tong University.}
\thanks{$^\dagger$Corresponding author email: wanghesheng@sjtu.edu.cn}
}

\begin{document}

\maketitle

\begin{abstract}
Robot manipulation policies often struggle to generalize beyond their demonstrations, even when new instructions involve familiar objects and behaviors.
When language and scenes are strongly correlated during training, a policy can learn a fixed visual--action mapping rather than respond to the requested behavior.
We investigate compositional reuse at two levels: within a subtask, combining familiar entities, action types, and action modifiers; and across subtasks, reusing learned subtasks in unseen long-horizon tasks.
We introduce CoMani, a benchmark with controlled splits for evaluating both capabilities.
Matched initial scenes and controlled changes to a single semantic factor encourage reliance on language rather than visual shortcuts.
We further propose GraphPoint, which connects semantic entity graphs to geometric control by predicting future gripper point trajectories and converting them into actions using robot geometry.
The framework organizes the gripper and objects by semantic roles and conditions their interactions on action types and modifiers, while predicted progress guides transitions during execution.
Experiments and ablations on CoMani validate the effectiveness of our method for instruction-dependent generalization at both levels.
Code will be released at \href{https://github.com/Gorgeousful/GraphPoint.git}{GraphPoint}.
\end{abstract}

\section{Introduction}

A robot that has learned to put a bowl on a plate should not be limited to repeating that exact behavior.
Given a new instruction, it should place the same bowl to the right of the plate, or sweep it there instead of lifting it.
Beyond such variations within a single interaction, the robot should combine familiar subtasks into longer instructions that were never demonstrated as a whole.
These capabilities require generalization both within individual subtasks and across subtasks in long-horizon execution.

Imitation learning has advanced through action chunking~\cite{zhao2023learning}, generative action prediction~\cite{chi2025diffusion}, and large-scale vision-language-action pretraining~\cite{intelligence2025pi_,kim2024openvla}.
Scene-point policies represent the environment with sparse 3D geometry~\cite{ze20243d,shridhar2023perceiver,goyal2023rvt}.
Entity-point policies instead focus on task-relevant objects and action keypoints~\cite{haldar2025point,haldar2026point,qi2025compose,zhu2023learning}.
These advances improve performance on demonstrated tasks, but do not necessarily produce behavior that responds to language in new settings~\cite{jiang2023vima}.
If a scene is repeatedly paired with the same motion, visual cues alone may provide an effective shortcut from observation to action.
The weakness becomes apparent when familiar entities are paired with a new action modifier or action type.
It becomes more pronounced in long-horizon execution, where each subtask must operate from the state produced by the preceding one~\cite{mees2022calvin,liu2023libero,zhang2025vlabench}.

\begin{figure}[t]
\centering
\includegraphics[width=\columnwidth]{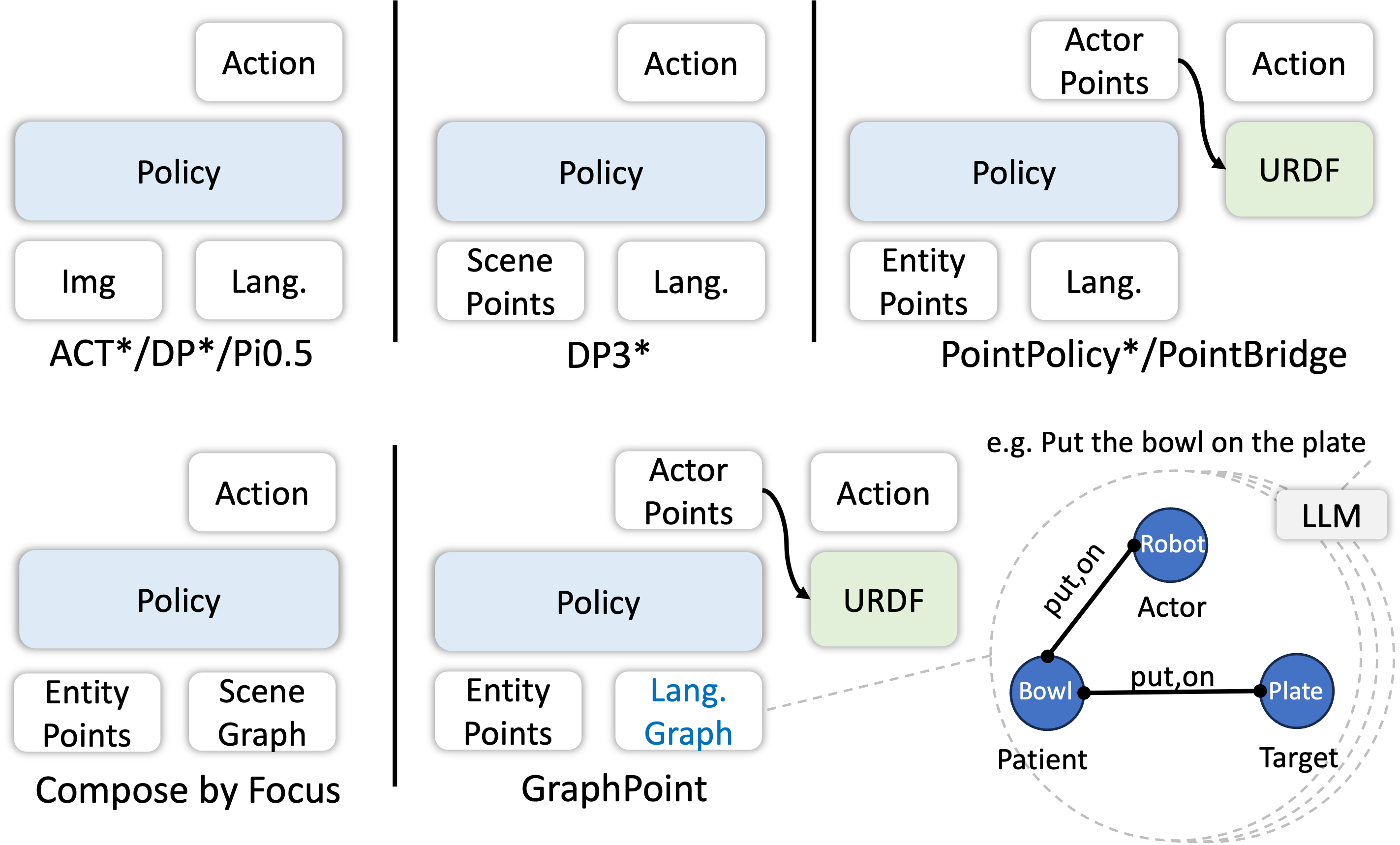}
\caption{Representative policies based on images, scene points, and entity points, 
with or without scene graphs; 
an asterisk marks baselines adapted with the frozen BGE language encoder~\cite{xiao2024c}.
GraphPoint uses a large language model (LLM) to parse the task instruction into a graph of role-specific entities, action type, and modifier, 
which, together with grounded entity points, conditions actor trajectory prediction.}
\label{fig:representation_comparison}
\end{figure}

We use compositional reuse to study and address these two forms of generalization.
Within an atomic interaction, a policy must apply familiar semantic factors in a new combination.
Across a sequence, it must reuse learned subtasks, manage their transitions, and recover from the state changes accumulated along the way.
Addressing both levels requires a controlled benchmark that reveals whether a policy indeed follows language rather than relying on a fixed mapping, and a framework that grounds task semantics even with the same observation.

We first introduce CoMani, a robosuite-based benchmark~\cite{zhu2020robosuite} organized around these two levels.
Its action-modifier and action-type suites use controlled in-distribution (ID) and out-of-distribution (OOD) splits 
while varying one semantic factor under matched initial scenes, 
making the instruction necessary for selecting the intended behavior.
The OOD splits test whether familiar factors can be recombined within an atomic interaction, 
while the sequence suite trains policies only on individual subtasks and evaluates their reuse in unseen long-horizon instructions.

Fig.~\ref{fig:representation_comparison} contrasts GraphPoint with representative image-, scene-point-, and entity-point-based policies.
We further propose GraphPoint, which encodes reliable manipulation priors directly into the policy structure and represents the current manipulation as a semantic entity graph, 
making the instructed relation explicit rather than implicit in correlated visual cues.
The graph keeps the gripper, manipulated object, and reference entity as separate role-specific nodes.
Action types and modifiers condition the exchange of geometric information between them.
Both the observed entity points and the predicted gripper trajectories are expressed relative to the current tool center point (TCP) rather than in absolute scene coordinates, 
reducing dependence on absolute scene placement.
Executable poses are then recovered from the predicted trajectories using robot geometry.
During long-horizon execution, the same atomic policy is reused across subtasks and predicted progress decides when to advance, so that learned subtasks can be recombined into instructions that were never demonstrated as a whole.

Our contributions are:
\begin{itemize}
    \item CoMani, a benchmark that evaluates controlled ID/OOD semantic recombination 
    within atomic interactions and subtask reuse across unseen long-horizon instructions.
    \item GraphPoint, a framework that uses role-structured semantic entity graphs to condition point-trajectory control and predicted progress to support transitions between subtasks.
    \item Extensive experiments on CoMani show strong gains over diverse image-, scene-point-, entity-point-, and graph-based baselines, 
    while ablation studies further support the framework design.
\end{itemize}

\section{Related Work}

\subsection{Visual and Scene-Point Representations}
Image-based policies such as Action Chunking with Transformers (ACT)~\cite{zhao2023learning}, Diffusion Policy~\cite{chi2025diffusion}, and recent vision-language-action (VLA) models~\cite{kim2024openvla} learn manipulation directly from visual observations, but task-relevant geometry remains implicit in image features.
To better capture spatial structure, PerAct~\cite{shridhar2023perceiver} and RVT~\cite{goyal2023rvt} exploit explicit 3D scene structure, while DP3~\cite{ze20243d} operates on compact scene point clouds for data-efficient visuomotor learning.
These approaches improve geometric generalization, yet encode the workspace as a largely unstructured scene, without distinguishing the semantic roles of task-relevant entities.
GraphPoint instead represents the scene as separate role-specific entities, each with its own geometric encoding.

\subsection{Entity-Centric and Graph Representations}
Recent works move beyond scene-level geometry by representing task-relevant entities with sparse points or graphs.
Point Policy~\cite{haldar2025point} uses semantic keypoints to unify observations and actions, while Point Bridge~\cite{haldar2026point} extends point-based representations to cross-domain (sim-to-real) transfer.
GraphMimic~\cite{chen2025graphmimic} represents objects and interactions as graphs and predicts future graph states to support policy learning from videos, while Compose by Focus (CbF)~\cite{qi2025compose} uses scene graphs to capture task-relevant objects and relations for robust atomic manipulation.
These methods demonstrate the value of entity-centric structure, but do not organize entities by manipulation roles or condition their interactions on compositional action semantics.
GraphPoint addresses this by coupling role-specific entity graphs with action-type- and modifier-conditioned point-trajectory control.

\subsection{Compositional and Long-Horizon Benchmarks}
VIMA~\cite{jiang2023vima} studies compositional generalization with multimodal prompts, and its four-level protocol includes a combinatorial level that recombines objects and textures seen separately during training, although its scenes are procedurally generated rather than matched across instructions.
Long-horizon domains such as CALVIN~\cite{mees2022calvin}, LIBERO-Long~\cite{liu2023libero}, and VLABench~\cite{zhang2025vlabench} evaluate multi-stage execution: CALVIN records unsegmented long-horizon play data and segments it afterwards, LIBERO-Long trains directly on its long-horizon tasks, and VLABench embeds several skills and sub-steps in a single instruction.
However, none of these settings isolates a single semantic change under matched initial scenes, or evaluates whether policies trained on separately recorded atomic subtasks, with no demonstration of a transition, can compose them into unseen long-horizon tasks.
CoMani targets both gaps through controlled semantic recombination and zero-shot subtask composition.

\section{Method}

\subsection{Overview}
\begin{figure*}[t]
\centering
\includegraphics[width=0.9\textwidth]{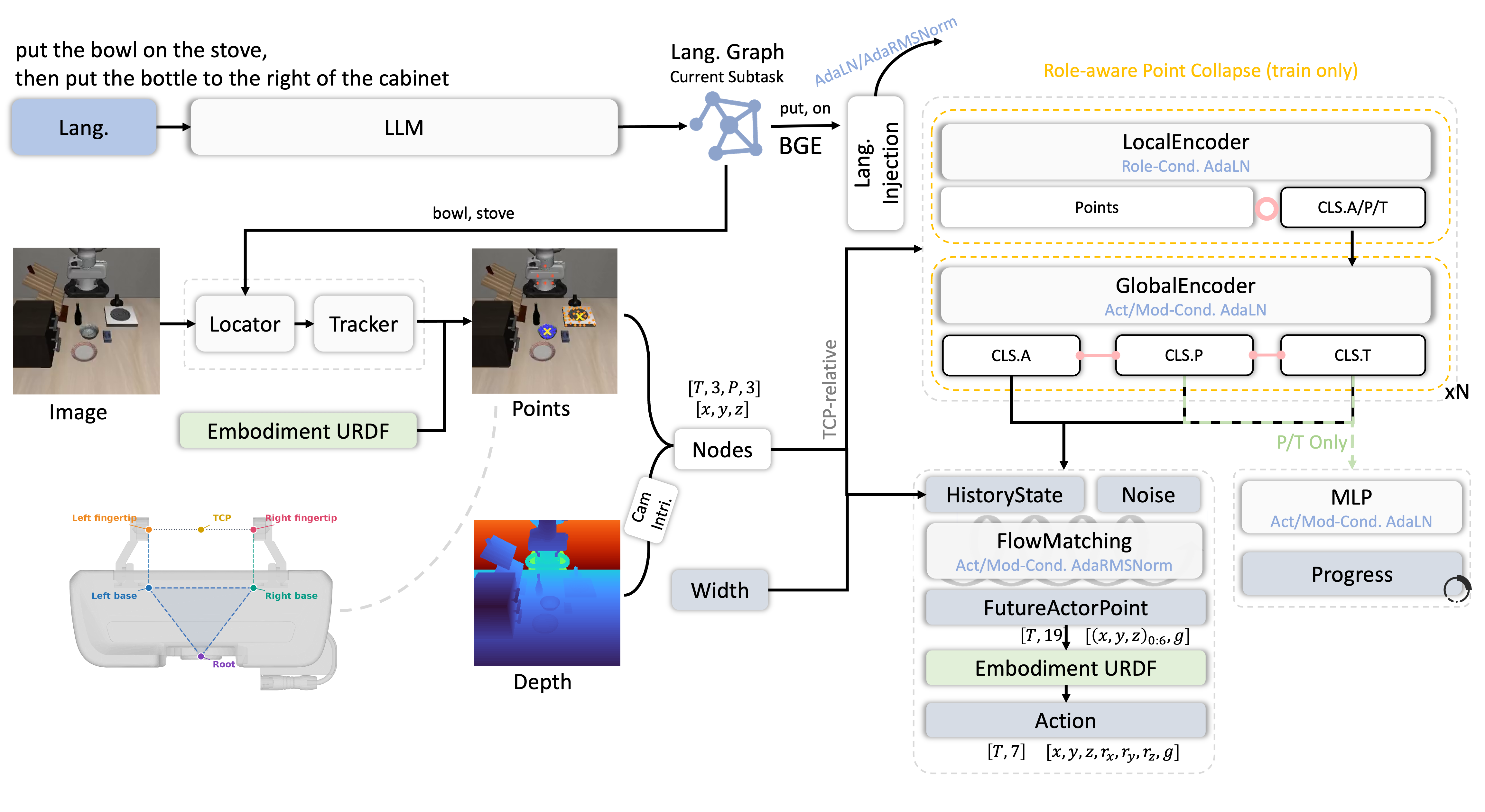}
\caption{GraphPoint framework.
Language analysis identifies ordered subtasks and their semantic roles, while visual localization and tracking ground the relevant entities as 3D points.
The policy alternates local entity encoding with global attention across entities, conditioned on roles, action types, and modifiers.
A flow decoder predicts future actor keypoints and gripper commands, and a progress head supports subtask transitions.
Robot geometry converts predicted points into executable gripper poses.}
\label{fig:framework}
\end{figure*}

GraphPoint represents manipulation as interactions among 3D point entities and predicts future gripper trajectories in the same geometric space.
As shown in Fig.~\ref{fig:framework}, language analysis, visual localization, 
and mask tracking jointly construct a language-conditioned entity graph, which is processed by a shared policy to predict both the gripper trajectory and subtask progress.
For multi-step instructions, the entity graph is instantiated for each subtask, while the predicted progress determines when the system advances to the next one.

\subsection{Task-to-Graph Construction}
\textbf{Semantic decomposition.} Similar to Fig.~\ref{fig:representation_comparison}, we use an LLM (e.g., GPT) to decompose an instruction into ordered subtask graphs. 
For example, ``put the left white bowl on the middle plate'' is parsed with the robotic gripper as the \emph{actor} $A$, 
the left white bowl as the \emph{patient} $P$, the middle plate as the \emph{target} $T$, 
\emph{put} as the action type, and \emph{on} as the modifier. 
For each subtask, entity roles are represented by learned embeddings $e_r$, $r\in\{A,P,T\}$, 
while entity descriptions, including spatial and appearance attributes, are used only for visual grounding.
The action type and modifier are encoded by a frozen BGE encoder~\cite{xiao2024c} and projected to $l_{\mathrm{act}}$ and $l_{\mathrm{mod}}$, respectively. 

\textbf{Grounded point entities.}
For \emph{patient} and \emph{target}, we use a spatially aware vision-language model (VLM) as the \emph{locator} to identify task-relevant objects or parts from entity descriptions extracted by the LLM.
A temporal segmentation model is then used as the \emph{tracker} to propagate their masks across frames.
In our implementation, LocateAnything~\cite{wang2026locateanything} serves as the locator and SAM~2~\cite{ravi2025sam} as the tracker.
From each tracked mask, we sample 32 points to represent the corresponding visual entity over time.
By default, we do not enforce point-level temporal correspondence for visual entity points with an additional point tracker such as TAPNext++~\cite{jung2026tapnextpp}, since point-level correspondence is not essential to our entity-level representation and point tracking can introduce unstable trajectories.

For the \emph{actor}, we use six gripper keypoints defined by robot geometry and instantiated at each frame from the robot state: 
the root and two finger bases are fixed anchors that form a rigid triangle for stable pose representation, 
while the two fingertips and the TCP are placed according to the observed finger opening and characterize the geometry relevant to contact.

\subsection{Semantic Conditioning}
All attention and feed-forward layers use condition-dependent normalization of $h$, with the scale and shift determined by $q$:
\begin{align}
    \operatorname{AdaLN}(h;q)&=(1+\gamma(q))\odot\operatorname{LN}(h)+\beta(q),\\
    \operatorname{AdaRMSNorm}(h;q)&=(1+\gamma(q))\odot\operatorname{RMS}(h)+\beta(q),\notag
\end{align}
where $\gamma$ and $\beta$ are broadcast over tokens; the encoder applies $\operatorname{AdaLN}$, and the decoder applies $\operatorname{AdaRMSNorm}$ with a third output $g$ gating each residual update.
The role embedding conditions the local entity blocks, $q=e_r$, assigning entity geometry to the actor, patient, or target.
The projected action type and modifier, $q=[l_{\mathrm{act}};l_{\mathrm{mod}}]$, condition the global blocks, the flow decoder, and the progress head, so that the same geometry is interpreted according to the instructed interaction, such as placing an object \emph{on} or to the \emph{right} of a target.

\subsection{Role-Structured Entity Graph Encoder}
\textbf{TCP-relative representation.}
The encoder observes the current frame and nine history frames.
All entity points are expressed relative to the current-frame TCP: we subtract its translation from every patient, target, and actor point across the observation window and the prediction horizon, so that coordinates share a single origin while keeping the camera axes.
This reduces dependence on absolute scene placement and keeps the input geometry directly comparable with the predicted trajectory.

\textbf{Local and global encoders.}
A shared point MLP embeds entity coordinates, with keypoint identity embeddings added for the actor.
Each entity at each observed time receives four learned CLS tokens to summarize its geometry.
The encoder alternates local encoding within each entity and global interaction among entity summaries:
\begin{equation}
\begin{aligned}
    (\widetilde C_r,F_r^{i+1})
       &=\operatorname{Local}_{i}(C_r^i,F_r^i;e_r),\\
    C_r^{i+1}
       &=\operatorname{Global}_{i}(\{\widetilde C_{r}\}_{r\in\{A,P,T\}};l_{\mathrm{act}},l_{\mathrm{mod}}).
\end{aligned}
\end{equation}
Here, $F_r$ and $C_r$ denote point features and CLS summaries, respectively, with observation-time indices omitted where they are not needed; we write $C_{t,r}$ for the summary of role $r$ at time $t$.
Local self-attention processes each entity and frame independently, conditioned on its role embedding $e_r$.
Global attention exchanges only CLS summaries across entities and observed times, conditioned on $l_{\mathrm{act}}$ and $l_{\mathrm{mod}}$.
Its connectivity follows the bidirectional chain $A\leftrightarrow P\leftrightarrow T$, with additional within-role temporal connections.
Direct actor--target attention is masked, so their interaction is mediated by the patient through successive layers.
After eight layers, the current-time CLS summaries form memory $M_t$, incorporating both geometry and interaction history.

\textbf{Role-aware point collapse (RPC).}
During training, RPC independently collapses each patient or target point set with probability $0.05$
to suppress shape-based shortcuts and encourage greater reliance on injected semantic conditions.
Patient points are replaced by the mean of the four nearest to the TCP, and target points by the entity centroid.

\subsection{Point-Trajectory Flow Policy}
We predict $H=10$ future steps, each containing six actor keypoints and a scalar gripper command:
\begin{equation}
    Y=\{(X^A_{t+h},g_{t+h})\}_{h=1}^{H}.
\end{equation}
Here, $X^A_{t+h}$ contains actor coordinates in the same TCP-relative frame, and $g_{t+h}$ is the gripper command.
Each step is represented by a 19-dimensional vector in point-wise $(x,y,z)$ order followed by the gripper command.
Representing actions as gripper points makes the predicted motion spatially comparable to the observed entities and encodes orientation through the arrangement of multiple points.

A six-layer Transformer models this trajectory with conditional flow matching~\cite{lipman2022flow}.
It embeds the observed actor history $U_t$ together with the noisy future trajectory $Z_\tau$ and predicts a velocity field:
\begin{equation}
    v_\theta=\operatorname{Decoder}_{\theta}(Z_\tau,U_t;M_t,l_{\mathrm{act}},l_{\mathrm{mod}},\tau).
\end{equation}
Each decoder layer uses self-attention to relate trajectory steps, cross-attention to retrieve entity geometry from $M_t$, and a feed-forward network to update features; a final projection maps the future tokens to point and gripper velocities.
Future tokens attend to both history and future, whereas history tokens attend only to history.
At inference, ten Euler steps integrate from noise at $\tau=1$ to a trajectory at $\tau=0$; we add the subtracted TCP translation back and recover the gripper pose command by least-squares rigid alignment of the six predicted actor keypoints to the reference keypoints given by robot geometry. 
With a 10-action chunk, the policy alone runs at 7.8\,Hz on a single RTX 5090, and the full pipeline including the locator and tracker runs at 2.1\,Hz.

% Submit the benchmark figure early so it appears beside the benchmark introduction.
\begin{figure*}[!t]
    \centering
    \includegraphics[width=\textwidth]{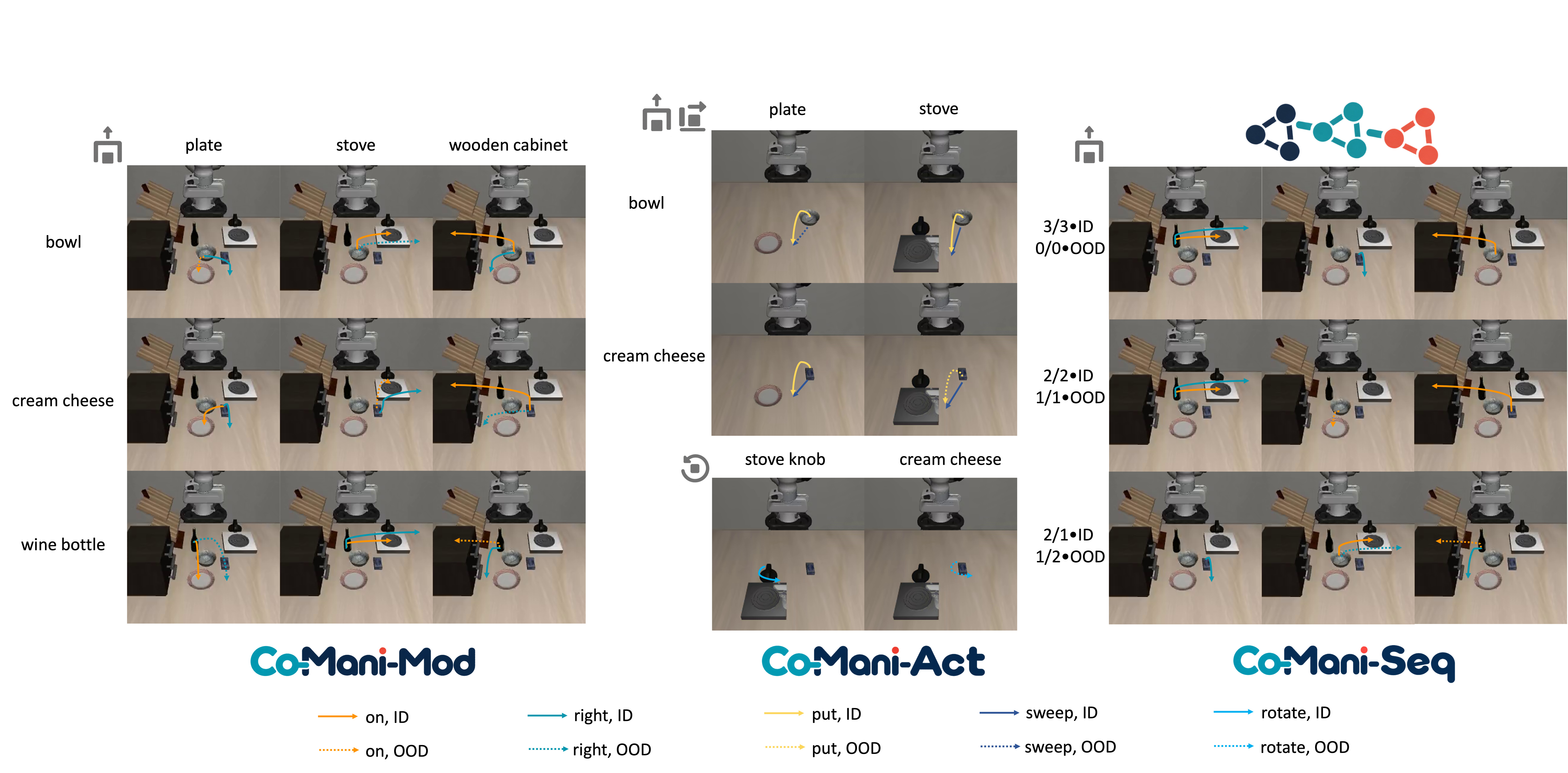}
    \caption{CoMani task construction and splits.
    Left: CoMani-Mod varies modifiers with a fixed put action.
    Center: CoMani-Act varies action types with a fixed right modifier.
    Right: CoMani-Seq composes atomic tasks from CoMani-Mod into six three-step sequences, shown in three rows with execution proceeding from left to right.}
    \label{fig:benchmark}
\end{figure*}

\subsection{Subtask Progress and Long-Horizon Composition}
The progress head estimates subtask completion from the patient--target relation:
\begin{equation}
s_t=\operatorname{Progress}_{\theta}(C_{t,P},C_{t,T};l_{\mathrm{act}},l_{\mathrm{mod}}).
\end{equation}
For actions without an explicit target, such as rotating a knob, the target CLS slot is replaced by the patient CLS summary from the subtask's initial frame, providing a reference for state change.
The head is a language-conditioned MLP with a sigmoid output.
At deployment, the planner advances when predicted progress exceeds 0.9 for two consecutive policy calls.
We denote this learned transition rule as self-switching (SS).
Environment-based switching (ES) instead uses an environment-provided subtask completion signal to separate execution performance from learned transition timing.
On a transition, the system releases and lifts the gripper, selects the next subtask's entities and language conditions, and resets policy history.
The same atomic policy is reused throughout the sequence.

\section{CoMani Benchmark}
\label{sec:benchmark}

\begin{table*}[!t]
\centering
\caption{CoMani task lists and ID/OOD splits.
In (c), ID/OOD gives the number of ID and OOD atomic subtasks; atomic task indices refer to (a), with OOD indices underlined.}
% IMPORTANT: Paper task indices differ from code/evaluation task IDs.
% Use expr/task_id_mapping.json when importing results; never match numeric indices directly.
\label{tab:comani_tasks}
\footnotesize
\setlength{\tabcolsep}{2pt}
\renewcommand{\arraystretch}{1.08}
\begin{minipage}[t]{0.445\textwidth}
\centering
\textbf{(a) CoMani-Mod}\par\smallskip
% Match the height of the stacked Act and Seq tables on the right.
\renewcommand{\arraystretch}{1.49}
\begin{tabular*}{\linewidth}{@{\extracolsep{\fill}}r|p{6.35cm}|l}
\noalign{\hrule height 1pt}
Task & Instruction & Split \\
\hline
0 & Put the bowl on the stove. & ID \\
1 & Put the bowl on the wooden cabinet. & ID \\
2 & Put the bowl to the right of the plate. & ID \\
3 & Put the bowl to the right of the wooden cabinet. & ID \\
4 & Put the cream cheese on the plate. & ID \\
5 & Put the cream cheese on the wooden cabinet. & ID \\
6 & Put the cream cheese to the right of the plate. & ID \\
7 & Put the cream cheese to the right of the stove. & ID \\
8 & Put the wine bottle on the plate. & ID \\
9 & Put the wine bottle on the stove. & ID \\
10 & Put the wine bottle to the right of the stove. & ID \\
11 & Put the wine bottle to the right of the wooden cabinet. & ID \\
\hline
12 & Put the bowl on the plate. & OOD \\
13 & Put the bowl to the right of the stove. & OOD \\
14 & Put the cream cheese on the stove. & OOD \\
15 & Put the cream cheese to the right of the wooden cabinet. & OOD \\
16 & Put the wine bottle on the wooden cabinet. & OOD \\
17 & Put the wine bottle to the right of the plate. & OOD \\
\noalign{\hrule height 1pt}
\end{tabular*}
\end{minipage}\hfill
\begin{minipage}[t]{0.535\textwidth}
\centering
\textbf{(b) CoMani-Act}\par\smallskip
\begin{tabular*}{\linewidth}{@{\extracolsep{\fill}}r|p{6.35cm}|l}
\noalign{\hrule height 1pt}
Task & Instruction & Split \\
\hline
0 & Put the bowl to the right of the plate. & ID \\
1 & Put the bowl to the right of the stove. & ID \\
2 & Put the cream cheese to the right of the plate. & ID \\
3 & Rotate the stove knob right. & ID \\
4 & Sweep the bowl to the right of the stove. & ID \\
5 & Sweep the cream cheese to the right of the plate. & ID \\
6 & Sweep the cream cheese to the right of the stove. & ID \\
\hline
7 & Put the cream cheese to the right of the stove. & OOD \\
8 & Rotate the cream cheese right. & OOD \\
9 & Sweep the bowl to the right of the plate. & OOD \\
\noalign{\hrule height 1pt}
\end{tabular*}
\par\smallskip
\centering
\textbf{(c) CoMani-Seq}\par\smallskip
\begin{tabular*}{\linewidth}{@{\extracolsep{\fill}}r|>{\centering\arraybackslash}m{0.85cm}|m{5.48cm}|>{\centering\arraybackslash}m{1.90cm}}
\noalign{\hrule height 1pt}
Task & ID/\linebreak OOD & Sequence & Atomic\linebreak tasks \\
\hline
0 & 3/0 & Wine bottle on stove $\to$ cream cheese right of plate $\to$ bowl on wooden cabinet & $9\to6\to1$ \\
1 & 3/0 & Wine bottle right of stove $\to$ cream cheese right of plate $\to$ bowl on wooden cabinet & $10\to6\to1$ \\
\hline
2 & 2/1 & Wine bottle on stove $\to$ bowl on plate $\to$ cream cheese on wooden cabinet & $9\to\underline{12}\to5$ \\
3 & 2/1 & Wine bottle right of stove $\to$ bowl on plate $\to$ cream cheese on wooden cabinet & $10\to\underline{12}\to5$ \\
\hline
4 & 2/1 & Cream cheese right of plate $\to$ bowl on stove $\to$ wine bottle on wooden cabinet & $6\to0\to\underline{16}$ \\
5 & 1/2 & Cream cheese right of plate $\to$ bowl right of stove $\to$ wine bottle on wooden cabinet & $6\to\underline{13}\to\underline{16}$ \\
\noalign{\hrule height 1pt}
\end{tabular*}
\end{minipage}
\end{table*}

CoMani comprises 3 suites testing modifier, action-type, and sequential composition, illustrated in Fig.~\ref{fig:benchmark} and listed in Table~\ref{tab:comani_tasks}.
We construct the dataset with 50 trajectories for each atomic task, generated using MimicGen~\cite{mandlekar2023mimicgen}.
Within each suite, initial scenes are matched wherever feasible to require instruction-dependent behavior and discourage fixed vision-to-action mappings.

% Figure corrections pending: in the CoMani-Mod legend, dotted yellow means
% on/OOD and solid blue means right/ID.

\textbf{CoMani-Mod.}
With the pick-and-place action \emph{put} fixed, this suite combines 3 objects, 3 targets, and the modifiers \emph{on} and \emph{right} into 18 tasks, and is evaluated by success rate (SR).
Fixing \emph{put} holds the action type constant, so two trials with the same initial scene differ only in the instructed relation, making failures easier to attribute to relation understanding rather than a different manipulation skill.
The split contains 12 ID tasks and 6 OOD tasks.
Each OOD object--target pair appears in the ID split with the opposite modifier, so all individual factors are familiar and success requires recombining them into an unseen composition rather than acquiring a new skill.

% Internal suite: 0902.
% Code ID-split IDs: 1,2,3,5,6,8,9,10,12,13,16,17.
% OOD IDs: 0,4,7,11,14,15.
% Held-out on: bowl/plate, cheese/stove, bottle/cabinet.
% Held-out right: bowl/stove, cheese/cabinet, bottle/plate.

\textbf{CoMani-Act.}
With \emph{right} fixed, this suite varies \emph{put}, \emph{sweep}, and \emph{rotate} across 10 tasks: 7 ID and 3 OOD.
Fixing the modifier isolates variation in the action type, so differences between tasks are attributable to the action type rather than to the relation or the goal.
For put/sweep, each OOD object--target pair appears in the ID split with the other action; for rotation, the ID task uses the stove knob and the OOD task uses cream cheese, testing whether an action transfers to a new entity.
Rotate is included as a unary action: unlike \emph{put} and \emph{sweep}, it has no target and operates on the manipulated entity alone.
Put/sweep pairs share initial states and final-position goals, but success additionally requires the execution to exhibit the instructed action.
We therefore report an \emph{action-consistent success rate} (ACSR), the fraction of trials that both satisfy the final goal and exhibit the instructed action type: lifting and transporting the object (put), pushing it along the support surface (sweep), or turning it or a knob in the instructed direction (rotate).
Reaching the same final position through a different action does not count.

% Internal suite: 0904.
% Code ID-split IDs: 0,1,2,5,7,8,9.
% Code OOD-split IDs: 3,4,6.
% Rotation details: dataset term "right" means counterclockwise about world +Z
% viewed from above.
% Knob: joint position >= 0.5 rad.
% Cheese: reset-relative signed rotation >= 0.5 rad, with tilt <= 15 degrees.

\textbf{CoMani-Seq.}
This suite tests whether policies learned from CoMani-Mod atomic demonstrations can execute 3-step instructions without sequence-level demonstrations.
We report $\mathrm{SR}_k$, the success rate of completing the first $k$ subtasks in order, for $k\in\{2,3\}$, together with a progress rate (PR) over the ordered subtask prefix.
It contains 6 sequences organized into 3 fixed-order pairs, each varying one modifier between \emph{on} and \emph{right}, without permuting subtasks, so a change of success within a pair can be traced to a single atomic factor rather than to reordering.
Three steps allow state and errors to accumulate across subtask boundaries while keeping evaluation comparable to the atomic suites.
The all-ID pair tests composition among familiar subtasks, while the remaining pairs test sequential execution with held-out atomic combinations.
The final pair increases the OOD count from 1 to 2 through a single modifier change, isolating the effect of additional atomic OOD.

% This suite tests whether policies learned from CoMani-Mod atomic demonstrations can execute 3-step instructions without sequence-level demonstrations, and reports SR together with a progress rate (PR) over the ordered subtask prefix.
% It contains 6 sequences organized into 3 fixed-order pairs, each varying one modifier between \emph{on} and \emph{right}, without permuting subtasks, so a change of success within a pair can be traced to a single atomic factor rather than to reordering.
% Three steps keep the horizon long enough for state and errors to accumulate, within an evaluation budget comparable to the atomic suites.
% The all-ID pair tests composition and transitions among familiar subtasks; the other pairs evaluate generalization to held-out atomic combinations, with the final pair increasing the OOD count from 1 to 2 via a single modifier change.

% Internal suite: libero_custom_0906.
% Evaluation order: 0,1,2,3,4,5.
% Figure pair order: tasks 2/3 (3 ID), 0/1 (2 ID + 1 OOD), 4/5 (mixed).
% Sequence initialization, scoring, and time budget are specified in Experiments.

\section{Experiments}

We organize the evaluation around four questions.
\begin{itemize}\itemsep0pt\topsep2pt\parsep0pt
\item \textbf{Q1}: Can GraphPoint generalize across modifiers?
\item \textbf{Q2}: Can GraphPoint generalize across actions?
\item \textbf{Q3}: Can GraphPoint compose atomic subtasks?
\item \textbf{Q4}: Which components matter?
\end{itemize}
Sec.~\ref{sec:exp_setup} describes the setup, and the following sections answer Q1--Q4 in turn.

\begin{table*}[!t]
\centering
\caption{CoMani-Mod success rate (SR).
Task indices follow Table~\ref{tab:comani_tasks}.}
\label{tab:modifier_comparison}
\fontsize{7}{8.4}\selectfont
\setlength{\tabcolsep}{1.5pt}
\renewcommand{\arraystretch}{1.15}
\setlength{\arrayrulewidth}{0.5pt}
\def\resultcolwidth{\dimexpr(\textwidth-2pt-1.7cm-42\tabcolsep-3\arrayrulewidth)/20\relax}
\begin{tabular}{p{1.7cm}|*{2}{>{\centering\arraybackslash}p{\resultcolwidth}}|*{12}{>{\centering\arraybackslash}p{\resultcolwidth}}|*{6}{>{\centering\arraybackslash}p{\resultcolwidth}}}
\noalign{\hrule height 1pt}
\multirow[c]{2}{*}{Method} & \multicolumn{2}{c|}{Mean} & \multicolumn{12}{c|}{ID} & \multicolumn{6}{c}{OOD} \\
\cline{2-21}
 & ID & OOD & 0 & 1 & 2 & 3 & 4 & 5 & 6 & 7 & 8 & 9 & 10 & 11 & 12 & 13 & 14 & 15 & 16 & 17 \\
\hline
% PointBridge sources: server 9996, point_bridge_points_custom0902-step_30000; ID libero_custom_0902-0912-025321-346811/result.json (103/360); OOD libero_custom_0902-0912-025322-202611/result.json (6/180).
% Verified against per-episode records and mapped through expr/task_id_mapping.json.
PointBridge & 28.6 & 3.3 & 0.0 & 36.7 & 60.0 & \textbf{100.0} & 0.0 & 0.0 & 83.3 & 63.3 & 0.0 & 0.0 & 0.0 & 0.0 & 20.0 & 0.0 & 0.0 & 0.0 & 0.0 & 0.0 \\
PointPolicy & 21.4 & 15.0 & 3.3 & 36.7 & 66.7 & 50.0 & 0.0 & 0.0 & 16.7 & 0.0 & 23.3 & 0.0 & 13.3 & 46.7 & 10.0 & 0.0 & 0.0 & 6.7 & 0.0 & 73.3 \\
% CbF Mod source: cbf_code_clip_custom0902_obs2_delta_canonical-step_30000, libero_custom_0902-oracle-retaskstructure-20260918/result.json; 540 completed episodes, 30 per task. Retrained and re-evaluated after the 2026-09-18 taskstructures fix, which made the atomic subtask string agree with the task string, so the language input now names the cream-cheese entity.
% Superseded pre-fix run: cbf_code_clip_custom0902_obs2_delta-step_30000, libero_custom_0902-oracle-0914-070206-099882 (code tasks 0--8) and -130851 (code tasks 9--17); ID 39.4 (142/360), OOD 6.7 (12/180). That run was train/inference consistent but labelled the cream-cheese entity "small blue box".
% Only 6 of the 18 tasks changed their subtask string (paper 4--7 and 14--15); their mean delta is +5.0pp versus +2.2pp for the other 12, so the difference between the two runs is dominated by retraining variance, not by the fix.
% Same checkpoint family and delta-action protocol as the CbF Act row; rates mapped through expr/task_id_mapping.json (Mod is a different task order from Act).
CbF & 46.1 & 2.8 & 80.0 & 70.0 & 96.7 & 66.7 & 0.0 & 43.3 & 3.3 & 56.7 & 20.0 & 76.7 & 20.0 & 20.0 & 3.3 & 3.3 & 6.7 & 3.3 & 0.0 & 0.0 \\
DP3 & 73.6 & 57.2 & 73.3 & \textbf{100.0} & \textbf{100.0} & \textbf{100.0} & 53.3 & \textbf{100.0} & 0.0 & 80.0 & 80.0 & 86.7 & 26.7 & 83.3 & 86.7 & 63.3 & 60.0 & \textbf{93.3} & 16.7 & 23.3 \\
ACT & 20.0 & 7.2 & 33.3 & 20.0 & 6.7 & 23.3 & 13.3 & 56.7 & 10.0 & 3.3 & 63.3 & 0.0 & 0.0 & 10.0 & 30.0 & 0.0 & 0.0 & 13.3 & 0.0 & 0.0 \\
DP & 40.0 & 10.0 & 43.3 & 36.7 & 16.7 & 26.7 & 50.0 & 66.7 & 20.0 & 33.3 & 80.0 & 16.7 & 36.7 & 53.3 & 30.0 & 0.0 & 0.0 & 30.0 & 0.0 & 0.0 \\
$\pi_{0.5}$ & 91.1 & 48.3 & \textbf{100.0} & \textbf{100.0} & 56.7 & \textbf{100.0} & 73.3 & \textbf{100.0} & 96.7 & \textbf{83.3} & \textbf{100.0} & \textbf{100.0} & 83.3 & \textbf{100.0} & 80.0 & 33.3 & 6.7 & 80.0 & 6.7 & \textbf{83.3} \\
\hline
\rowcolor{rowblue}
% GraphPoint OOD columns (paper tasks 0,4,7,11,14,15): 0902 OOD rerun of the same step_30000 checkpoint, 30 episodes per task, predicted switching (libero_custom_0902-0916-011040-742207 and -785922). ID columns are unchanged.
GraphPoint & \textbf{96.4} & \textbf{80.0} & \textbf{100.0} & \textbf{100.0} & \textbf{100.0} & 86.7 & \textbf{93.3} & \textbf{100.0} & \textbf{100.0} & 80.0 & \textbf{100.0} & \textbf{100.0} & \textbf{96.7} & \textbf{100.0} & \textbf{93.3} & \textbf{83.3} & \textbf{73.3} & 80.0 & \textbf{80.0} & 70.0 \\
\noalign{\hrule height 1pt}
\end{tabular}
\end{table*}

\begin{table*}[!t]
\centering
\caption{CoMani-Act action-consistent success rate (ACSR).
Task indices follow Table~\ref{tab:comani_tasks}.}
\label{tab:action_comparison}
\fontsize{7}{8.4}\selectfont
\setlength{\tabcolsep}{1.5pt}
\renewcommand{\arraystretch}{1.15}
\setlength{\arrayrulewidth}{0.5pt}
\def\resultcolwidth{\dimexpr(\textwidth-2pt-1.7cm-26\tabcolsep-3\arrayrulewidth)/12\relax}
\begin{tabular}{p{1.7cm}|*{2}{>{\centering\arraybackslash}p{\resultcolwidth}}|*{7}{>{\centering\arraybackslash}p{\resultcolwidth}}|*{3}{>{\centering\arraybackslash}p{\resultcolwidth}}}
\noalign{\hrule height 1pt}
\multirow[c]{2}{*}{Method} & \multicolumn{2}{c|}{Mean} & \multicolumn{7}{c|}{ID} & \multicolumn{3}{c}{OOD} \\
\cline{2-13}
 & ID & OOD & 0 & 1 & 2 & 3 & 4 & 5 & 6 & 7 & 8 & 9 \\
\hline
PointPolicy & 80.5 & 4.4 & \textbf{100.0} & 73.3 & 23.3 & \textbf{100.0} & \textbf{100.0} & 76.7 & 90.0 & 13.3 & 0.0 & 0.0 \\
CbF & 90.5 & 48.9 & 96.7 & 96.7 & 90.0 & 96.7 & \textbf{100.0} & 76.7 & 76.7 & 46.7 & 0.0 & \textbf{100.0} \\
DP3 & 83.3 & 60.0 & 80.0 & 93.3 & \textbf{96.7} & \textbf{100.0} & \textbf{100.0} & 86.7 & 26.7 & \textbf{80.0} & 0.0 & \textbf{100.0} \\
$\pi_{0.5}$ & \textbf{97.6} & 57.8 & \textbf{100.0} & 96.7 & 86.7 & \textbf{100.0} & \textbf{100.0} & \textbf{100.0} & \textbf{100.0} & 76.7 & 0.0 & 96.7 \\
\hline
% GraphPoint OOD-7 (paper 7 / code task 3): SAM3 node-segmenter re-run 12/30 = 40.0%; the original SAM2 value was 8/30 = 26.7%.
\rowcolor{rowblue}
GraphPoint & 90.5 & \textbf{68.9} & 50.0 & \textbf{100.0} & 83.3 & \textbf{100.0} & \textbf{100.0} & \textbf{100.0} & \textbf{100.0} & 40.0 & \textbf{76.7} & 90.0 \\
\noalign{\hrule height 1pt}
\end{tabular}
\end{table*}

\begin{table*}[!t]
\centering
\caption{CoMani-Seq $\mathrm{SR}_2$, $\mathrm{SR}_3$, and progress rate (PR). ES uses oracle goal signals; SS uses predicted progress.}
\label{tab:sequence_comparison}
\fontsize{6}{8.4}\selectfont
\setlength{\tabcolsep}{0.5pt}
\renewcommand{\arraystretch}{1.15}
\setlength{\arrayrulewidth}{0.5pt}
\def\resultcolwidth{\dimexpr(\textwidth-2pt-1.0cm-0.7cm-64\tabcolsep-11\arrayrulewidth)/30\relax}
\begin{tabular}{p{1.0cm}|>{\centering\arraybackslash}p{0.7cm}|*{3}{>{\centering\arraybackslash}p{\resultcolwidth}}|*{8}{*{3}{>{\centering\arraybackslash}p{\resultcolwidth}}|}*{3}{>{\centering\arraybackslash}p{\resultcolwidth}}}
\noalign{\hrule height 1pt}
\multirow[c]{3}{*}{Method} & \multirow[c]{3}{*}{Switch} & \multicolumn{12}{c|}{Mean} & \multicolumn{6}{c|}{3/0 ID + 3/0 OOD} & \multicolumn{6}{c|}{2/1 ID + 2/1 OOD} & \multicolumn{6}{c}{2/1 ID + 1/2 OOD} \\
\cline{3-32}
 & & \multicolumn{3}{c|}{All} & \multicolumn{3}{c|}{0 OOD} & \multicolumn{3}{c|}{1 OOD} & \multicolumn{3}{c|}{2 OOD} & \multicolumn{3}{c|}{0} & \multicolumn{3}{c|}{1} & \multicolumn{3}{c|}{2} & \multicolumn{3}{c|}{3} & \multicolumn{3}{c|}{4} & \multicolumn{3}{c}{5} \\
\cline{3-32}
 & & SR$_2$ & SR$_3$ & PR & SR$_2$ & SR$_3$ & PR & SR$_2$ & SR$_3$ & PR & SR$_2$ & SR$_3$ & PR & SR$_2$ & SR$_3$ & PR & SR$_2$ & SR$_3$ & PR & SR$_2$ & SR$_3$ & PR & SR$_2$ & SR$_3$ & PR & SR$_2$ & SR$_3$ & PR & SR$_2$ & SR$_3$ & PR \\
\hline
% PointPolicy Seq source: point_policy_custom0902/checkpoints/step_30000.pt (no sequence-level training), campaign seq20ep_tableIV_retaskstructure_20260918/PP_ES_full/point_policy_custom0902-step_30000/libero_custom_0906-oracle-full-20260918/result.json; 20 episodes per task, 120 total, oracle switching, absolute actions; re-run after the 2026-09-18 taskstructures fix. PR is mapped through expr/task_id_mapping.json (0 OOD = paper 0--1, 1 OOD = paper 2--4, 2 OOD = paper 5).
% Superseded pre-fix run: seq20ep_tableIV_20260914/merged/PP_ES/result.json (All PR 0.28, 0 OOD 0.83, 1 OOD 0.00, paper 3 PR 0.00). PP is conditioned on the current subtask string (session.current_subtask), which carried the node-name pollution on code tasks 4 and 5 (subtask 0) and code task 3 (subtask 1). The new/old difference is one rollout on code task 1 (paper 3), whose subtask texts and node list are identical in both runs, so it is re-run noise rather than an effect of the fix. SR>=2 and SR>=3 are 0.0 on every task.
PointPolicy & ES & 0.0 & 0.0 & 0.6 & 0.0 & 0.0 & 0.8 & 0.0 & 0.0 & 0.6 & 0.0 & 0.0 & 0.0 & 0.0 & 0.0 & 0.0 & 0.0 & 0.0 & 1.7 & 0.0 & 0.0 & 0.0 & 0.0 & 0.0 & 1.7 & 0.0 & 0.0 & 0.0 & 0.0 & 0.0 & 0.0 \\
% CbF Seq source: cbf_code_clip_custom0902_obs2_delta_canonical-step_30000 (same CbF checkpoint as the CoMani-Mod row), campaign seq20ep_tableIV_retaskstructure_20260918/CbF_ES_full; 20 episodes per task, 120 total, oracle switching, delta actions; retrained and re-evaluated after the 2026-09-18 taskstructures fix.
% Superseded pre-fix run: cbf_code_clip_custom0902_obs2_delta-step_30000, campaign seq20ep_CbF_ES_20260914 (All SR>=2 2.5, SR>=3 0.0, PR 14.2). That run was train/inference consistent but its subtask text said "small blue box" for the cream-cheese entity; its three SR>=2 rollouts did not reproduce in five repeats each under the identical protocol, so they are not carried over here.
% Values from CbF_ES_full/cbf_code_clip_custom0902_obs2_delta_canonical-step_30000/libero_custom_0906-oracle-full-20260918/result.json mapped through expr/task_id_mapping.json; grouping follows the other rows (0 OOD = tasks 0--1, 1 OOD = tasks 2--4, 2 OOD = task 5).
CbF & ES & 0.0 & 0.0 & 10.3 & 0.0 & 0.0 & 16.7 & 0.0 & 0.0 & 9.4 & 0.0 & 0.0 & 0.0 & 0.0 & 0.0 & 21.7 & 0.0 & 0.0 & 11.7 & 0.0 & 0.0 & 18.3 & 0.0 & 0.0 & 10.0 & 0.0 & 0.0 & 0.0 & 0.0 & 0.0 & 0.0 \\
% DP3 Seq source: dp3_bge_custom0902_obs2/checkpoints/step_30000.pt (no sequence-level training), campaign seq20ep_tableIV_retaskstructure_20260918/DP3_ES_full; 20 episodes per task, 120 total, oracle switching, delta actions, re-run after the 2026-09-18 taskstructures fix.
% The single DP3 rollout that reached full ordered progress (code task 0 = paper seq 2, init state 17) is counted as a
% failure: five repeats of that episode under the identical protocol (same checkpoint, oracle switching, 1500 steps,
% 20 Hz, num_steps_wait 30, seed 42, delta actions, one server process so the diffusion noise draws are independent)
% all failed at the second subtask. Campaign
% dp3_bge_custom0902_obs2-step_30000/libero_custom_0906-dp3ep17-repro5x-20260918: 0/5 success, prefix 1/3 in every
% repeat. PR is left unchanged because it reports the partial progress actually observed in the original 120 rollouts.
DP3 & ES & 0.0 & 0.0 & 12.8 & 0.0 & 0.0 & 16.7 & 0.0 & 0.0 & 13.9 & 0.0 & 0.0 & 1.7 & 0.0 & 0.0 & 26.7 & 0.0 & 0.0 & 6.7 & 0.0 & 0.0 & 30.0 & 0.0 & 0.0 & 11.7 & 0.0 & 0.0 & 0.0 & 0.0 & 0.0 & 1.7 \\
% PI05 Seq source: pi05_libero_custom0902_low_mem_finetune/custom0902_id10ep/29999 (no sequence-level training), campaign seq20ep_tableIV_retaskstructure_20260918/PI05_ES_full; 20 episodes per task, 120 total, oracle switching, delta actions, re-run after the 2026-09-18 taskstructures fix.
% SR_2 and SR_3 are 0.0 on every task; PR is mapped through expr/task_id_mapping.json (0 OOD = paper 0--1, 1 OOD = paper 2--4, 2 OOD = paper 5).
$\pi_{0.5}$ & ES & 0.0 & 0.0 & 32.2 & 0.0 & 0.0 & 32.5 & 0.0 & 0.0 & 31.7 & 0.0 & 0.0 & 33.3 & 0.0 & 0.0 & 33.3 & 0.0 & 0.0 & 31.7 & 0.0 & 0.0 & 33.3 & 0.0 & 0.0 & 30.0 & 0.0 & 0.0 & 31.7 & 0.0 & 0.0 & 33.3 \\
\hline
\rowcolor{rowblue}
GraphPoint & ES & 82.5 & \textbf{50.8} & \textbf{77.5} & \textbf{100.0} & \textbf{82.5} & \textbf{95.8} & 95.0 & \textbf{45.0} & \textbf{78.3} & 10.0 & 5.0 & 38.3 & \textbf{100.0} & 80.0 & \textbf{95.0} & \textbf{100.0} & \textbf{85.0} & \textbf{96.7} & \textbf{100.0} & \textbf{35.0} & 78.3 & 85.0 & \textbf{20.0} & 63.3 & \textbf{100.0} & \textbf{80.0} & \textbf{93.3} & 10.0 & 5.0 & 38.3 \\
\rowcolor{rowblue}
GraphPoint & SS & \textbf{86.7} & 38.3 & 76.1 & 95.0 & 67.5 & 87.5 & \textbf{96.7} & 26.7 & 76.7 & \textbf{40.0} & \textbf{15.0} & \textbf{51.7} & 95.0 & \textbf{85.0} & 91.7 & 95.0 & 50.0 & 83.3 & \textbf{100.0} & 30.0 & \textbf{80.0} & \textbf{90.0} & 10.0 & \textbf{68.3} & \textbf{100.0} & 40.0 & 81.7 & \textbf{40.0} & \textbf{15.0} & \textbf{51.7} \\
\noalign{\hrule height 1pt}
\end{tabular}
\end{table*}
% SR>=2 (prefix SR) = fraction of episodes with progress >= 2/3, i.e. the first two of three subtasks completed.
% Per-model SR>=2 in the All column: PP_ES 0.0, CbF_ES 0.0, DP3_ES 0.0, PI05_ES 0.0, GP_ES 82.5, GP_SS 86.7 (percent).
% Strict ordered prefix (completed_subtasks[:2] == [1, 2]) gives identical rates, so PR >= 2/3 is exact on this data.
% SR>=3 is the previous SR column; sources and ID/OOD grouping are unchanged.
% SR>=3 on the two GraphPoint rows is the environment final-goal conjunction (the previous SR column), not the
% ordered-prefix count that SR>=2 uses: GP_ES is 61/120 = 50.8 versus 63/120 = 52.5, and GP_SS is 46/120 = 38.3
% versus 53/120 = 44.2. The two definitions coincide for PP 0/0 and CbF 0/0 and for
% DP3 0/0 (its single full-progress rollout is discounted, see above) and PI05 0/0, so only the GraphPoint rows mix them; the gap is 2 (ES) and 7 (SS) rollouts that complete
% all three subtasks in order but leave a previously met goal false at the end. Under ordered prefix the cells
% would be GP_ES col0/col1 85.0/90.0 and 0 OOD 87.5, GP_SS col1--col4 60.0/40.0/20.0/45.0, 0 OOD 72.5, 1 OOD 35.0.
\begin{table*}[!tp]
\centering
\caption{CoMani-Mod ablation success rate (SR).
Task indices follow Table~\ref{tab:comani_tasks}.}
\label{tab:modifier_ablation}
\fontsize{7}{8.4}\selectfont
\setlength{\tabcolsep}{1.5pt}
\renewcommand{\arraystretch}{1.15}
\setlength{\arrayrulewidth}{0.5pt}
\def\resultcolwidth{\dimexpr(\textwidth-2pt-3.0cm-42\tabcolsep-3\arrayrulewidth)/20\relax}
\begin{tabular}{p{3.0cm}|*{2}{>{\centering\arraybackslash}p{\resultcolwidth}}|*{12}{>{\centering\arraybackslash}p{\resultcolwidth}}|*{6}{>{\centering\arraybackslash}p{\resultcolwidth}}}
\noalign{\hrule height 1pt}
\multirow[c]{2}{*}{Variant} & \multicolumn{2}{c|}{Mean} & \multicolumn{12}{c|}{ID} & \multicolumn{6}{c}{OOD} \\
\cline{2-21}
 & ID & OOD & 0 & 1 & 2 & 3 & 4 & 5 & 6 & 7 & 8 & 9 & 10 & 11 & 12 & 13 & 14 & 15 & 16 & 17 \\
\hline
w/o RPC & 90.6 & 44.4 & \textbf{100.0} & 96.7 & \textbf{100.0} & \textbf{100.0} & 66.7 & 96.7 & 96.7 & 83.3 & 93.3 & \textbf{100.0} & 70.0 & 83.3 & 20.0 & 90.0 & \textbf{80.0} & 6.7 & 10.0 & 60.0 \\
Full connectivity & 95.3 & 63.9 & 96.7 & 96.7 & \textbf{100.0} & \textbf{100.0} & \textbf{100.0} & 83.3 & \textbf{100.0} & 86.7 & \textbf{100.0} & \textbf{100.0} & 80.0 & \textbf{100.0} & 66.7 & 86.7 & 70.0 & 50.0 & 36.7 & 73.3 \\
Merged entity point set & 82.8 & 27.8 & 93.3 & 96.7 & \textbf{100.0} & \textbf{100.0} & 93.3 & 26.7 & 83.3 & 56.7 & \textbf{100.0} & 96.7 & 50.0 & 96.7 & 60.0 & 0.0 & 36.7 & 3.3 & 3.3 & 63.3 \\
% Raw language conditioning: retrained as 0906-pointdropknn005-basetcpfinger-cls4-rawlang-current-progress-sam-custom0902-ep10-rel-canonical/checkpoints/step_30000.pt (same training setup as the GraphPoint row, semantic_injection='raw_language'), then re-evaluated by outputs/0906-pointdropknn005-basetcpfinger-cls4-rawlang-current-progress-sam-custom0902-ep10-rel-canonical-step_30000/libero_custom_0902-predicted-retaskstructure-20260918/result.json; CoMani-Mod, 18 tasks x 30 episodes = 540, predicted switching, absolute actions; the raw subtask instruction is the only language input, with the role and modifier slots left empty.
% Superseded pre-fix runs (raw subtask text polluted by node names): libero_custom_0902-0915-152658-193325, -0915-152658-163213, and the OOD re-measuring runs -0916-023044-193738/-115823.
Raw language conditioning & 96.1 & 77.2 & 93.3 & \textbf{100.0} & \textbf{100.0} & \textbf{100.0} & 90.0 & 90.0 & \textbf{100.0} & \textbf{96.7} & \textbf{100.0} & 96.7 & 93.3 & 93.3 & \textbf{100.0} & \textbf{100.0} & 50.0 & 53.3 & 60.0 & \textbf{100.0} \\
% w/o language conditioning source: 0906-pointdropknn005-basetcpfinger-cls4-nulllang-current-progress-sam-custom0902-ep10-rel-step_30000 (same training setup as the GraphPoint row, semantic_injection='null'); runs libero_custom_0902-0915-154908-783050 (ID tasks 1,2,3,5,6), -154908-779063 (ID tasks 8,9,10,12,13), -162536-393388 (ID task 16), -164031-449221 (ID task 17), -154908-742548 (OOD tasks 0,4,7) and -154908-848717 (OOD tasks 11,14,15); 30 episodes per task, predicted switching on a second host; both semantic slots are left empty.
w/o language conditioning & 60.3 & 8.3 & 30.0 & 83.3 & 100.0 & 23.3 & 3.3 & 26.7 & 100.0 & 80.0 & 96.7 & 53.3 & 46.7 & 80.0 & 0.0 & 40.0 & 3.3 & 0.0 & 6.7 & 0.0 \\
Absolute point coordinates & 92.5 & 25.0 & 90.0 & \textbf{100.0} & \textbf{100.0} & 96.7 & 86.7 & \textbf{100.0} & 96.7 & 46.7 & \textbf{100.0} & \textbf{100.0} & 93.3 & \textbf{100.0} & 3.3 & 63.3 & 3.3 & 6.7 & 6.7 & 66.7 \\
Delta-action output & 95.0 & 73.3 & 96.7 & \textbf{100.0} & \textbf{100.0} & \textbf{100.0} & \textbf{100.0} & 96.7 & 80.0 & 86.7 & \textbf{100.0} & \textbf{100.0} & 80.0 & \textbf{100.0} & 96.7 & 86.7 & 76.7 & 76.7 & 50.0 & 53.3 \\
\hline
\rowcolor{rowblue}
% GraphPoint OOD columns: same 0902 OOD rerun as Table~\ref{tab:modifier_comparison}, 30 episodes per task.
GraphPoint & \textbf{96.4} & \textbf{80.0} & \textbf{100.0} & \textbf{100.0} & \textbf{100.0} & 86.7 & 93.3 & \textbf{100.0} & \textbf{100.0} & 80.0 & \textbf{100.0} & \textbf{100.0} & \textbf{96.7} & \textbf{100.0} & 93.3 & 83.3 & 73.3 & \textbf{80.0} & \textbf{80.0} & 70.0 \\
\noalign{\hrule height 1pt}
\end{tabular}
\end{table*}

\begin{figure*}[!t]
\centering
\includegraphics[width=\textwidth]{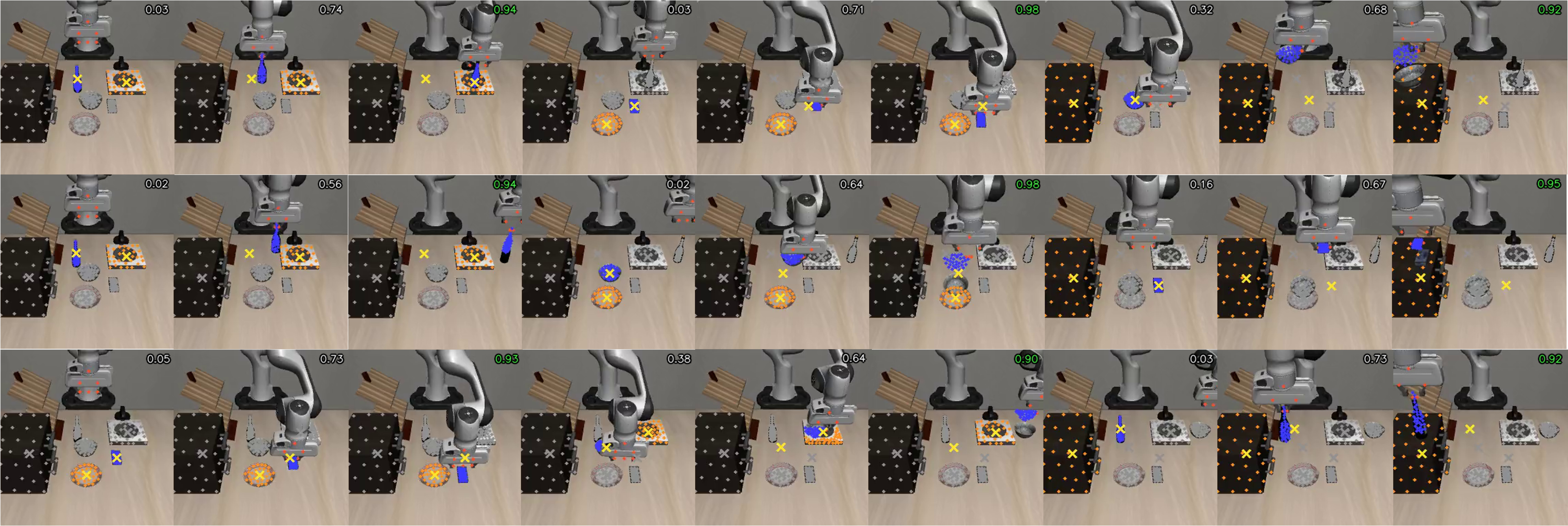}
\caption{Successful GraphPoint SS rollouts on CoMani-Seq.
Rows show sequences 0, 3, and 5 of Table~\ref{tab:comani_tasks}, which contain 0, 1, and 2 OOD subtasks, respectively, from top to bottom.
Each row proceeds from left to right, with 3 frames per subtask showing its start, execution, and completion.
Upper-right values show predicted subtask progress.}
\label{fig:sequence_rollouts}
\end{figure*}

\subsection{Experimental Setup}\label{sec:exp_setup}
\textbf{Baselines and adaptation.}
We compare GraphPoint with ACT~\cite{zhao2023learning}, DP~\cite{chi2025diffusion}, $\pi_{0.5}$~\cite{intelligence2025pi_}, DP3~\cite{ze20243d}, Point Policy~\cite{haldar2025point}, Point Bridge~\cite{haldar2026point}, and CbF~\cite{qi2025compose} on CoMani-Mod.
On CoMani-Act and CoMani-Seq we report one baseline per representation family, keeping the member that is strongest on held-out modifiers---$\pi_{0.5}$ (image-based), DP3 (scene points), Point Policy (entity points), and CbF (scene graphs).
% Family-selection rationale (the sentence above is kept as written): pi0.5, DP3, and Point Policy are the strongest
% held-out-modifier members of their families. For scene graphs, the 2026-09-18 taskstructures fix moved CbF's
% CoMani-Mod OOD mean from 6.7 to 2.8 (5/180 rollouts) while Point Bridge sits at 3.3 (6/180): one episode apart,
% i.e. statistically indistinguishable on 180 held-out rollouts. CbF is retained because on the same suite it is far
% stronger overall (ID 46.1 vs 28.6) and it is the family member carried into CoMani-Act and CoMani-Seq.
CbF, Point Bridge, and $\pi_{0.5}$ keep their original designs and native language conditioning, with no added language encoder; $\pi_{0.5}$ is fine-tuned with LoRA.
We adapt ACT, DP, DP3, and Point Policy to multi-task instructions using frozen BGE embeddings of the full instruction, without role/action/modifier decomposition.
ACT projects this embedding into an additional Transformer encoder token; Point Policy appends a projected language token to its point-token sequence and, following the original method, also uses point tracking~\cite{jung2026tapnextpp} after SAM~2.
DP and DP3 combine language and observation features through their existing FiLM~\cite{perez2018film} conditioning paths.
For CoMani-Seq, atomic policies are reused without sequence-level training, receiving the active subtask instruction under a shared ES protocol; GraphPoint is also evaluated with SS.

\textbf{Training and evaluation.}
All models are trained for 30,000 steps with batch size 64 and AdamW; OOD tasks are excluded from training, and we do not use an exponential moving average (EMA).
We execute actions at 20\,Hz over 30 trials per task on CoMani-Mod and CoMani-Act, and 20 trials per sequence on CoMani-Seq, using the splits in Sec.~\ref{sec:benchmark}.
For every corresponding rollout, all methods use the same scene initialization.
We report the suite-specific metrics of Sec.~\ref{sec:benchmark} as task means and ID/OOD aggregates.

% IMPORTANT: Map code IDs through expr/task_id_mapping.json before entering results.
% Existing numeric result columns retain their original task correspondence.
% Sources: expr/1.csv (Modifier comparison and ablations), expr/2.csv (Action).
% PI05 CoMani-Mod source: openpi/data/libero/pi05_custom0902_step29999/{id,ood}, 30 trials/task.
% Aggregate rates are copied from CSV SR-ID/SR-OOD, avoiding re-averaging rounded task percentages.

\subsection{Can GraphPoint Generalize Across Modifiers?}\label{sec:exp_mod}
Table~\ref{tab:modifier_comparison} shows GraphPoint as the most accurate method on both seen and held-out modifiers, and the only one that stays above 60\% on every held-out task.
The baselines split into two regimes by ID success: ACT, DP, Point Policy, Point Bridge, and CbF acquire the seen modifiers only partially, so their held-out scores measure missing acquisition rather than a failure to generalize, whereas $\pi_{0.5}$ and DP3 acquire them but do not sustain that level once the modifier is held out.
This suggests that these policies may bind the modifier to visual context, limiting transfer when a familiar object is paired with a held-out relation.
GraphPoint instead conditions on the modifier through the language and role structure, keeping a high success floor across held-out tasks and improving modifier generalization, though a residual gap remains.

\subsection{Can GraphPoint Generalize Across Actions?}\label{sec:exp_act}
Table~\ref{tab:action_comparison} shows GraphPoint attaining the highest held-out ACSR while retaining most of its seen-action accuracy, 
whereas the two methods that lead on seen actions, $\pi_{0.5}$ and CbF, show substantial degradation on the held-out tasks.
The three held-out tasks cover two cases: recombining a familiar object--target pair under the other action, and applying rotation to an entity that was never rotated.
GraphPoint handles both, and is the only method with action-consistent success on the rotation task, where DP3, the strongest baseline on the recombinations, records none.
This demonstrates that our method not only executes the instructed action on demonstrated entities, but can also transfer it to a held-out entity.
% PointPolicy Act sources: point_policy_custom0904-step_30000/libero_custom_0904-0913-131340-809426 and libero_custom_0904-0913-131421-638036; 300 unique completed episodes; author action review excludes 13 terminal successes for code task 3 (paper task 7), leaving 4/30.
% Raw final_goals records are preserved.
% Paper order: 0,1,2,5,7,8,9,3,4,6.
% The author confirmed that the reported CoMani-Act rates include action-process checks.

\subsection{Can GraphPoint Compose Atomic Subtasks?}\label{sec:exp_comp}
Table~\ref{tab:sequence_comparison} evaluates three-step instructions that were never demonstrated as a whole, with each step receiving only its own subtask instruction.
Although the baselines make partial progress, none of them completes even the first two subtasks (0.0\% $\mathrm{SR}_2$ for every baseline), and none completes a full three-step sequence.
In contrast, GraphPoint achieves 82.5\% $\mathrm{SR}_2$ and 50.8\% $\mathrm{SR}_3$ with ES.
Its transition signal is learned as well. Self-switching advances purely on predicted progress, with no environment goal signal, and matches ES in progress rate while trading some full-sequence success.
This shows that GraphPoint can compose subtasks learned in isolation into long-horizon instructions it was never trained on.
Fig.~\ref{fig:sequence_rollouts} visualizes successful self-switching executions from all three OOD counts, with the entity points and predicted progress overlaid on the frames.

\subsection{Which Components Matter?}\label{sec:exp_abl}
Table~\ref{tab:modifier_ablation} examines seven GraphPoint variants on CoMani-Mod, which remove RPC, replace the bidirectional chain connectivity of the global encoder with full connectivity, merge all entity points into a single set, vary the language conditioning, use absolute instead of TCP-relative point coordinates, and predict Cartesian delta actions instead of point trajectories.

\textbf{Geometry and graph structure.}
Removing RPC keeps seen-modifier performance close to the full model but substantially reduces held-out performance, 
supporting RPC as a regularizer for semantic recombination.
Full connectivity likewise preserves seen-modifier performance while lowering held-out performance, supporting patient-mediated interaction as a useful inductive bias rather than a capacity limit.
Merging the entity points into a single set is the most damaging structural variant, removing role embeddings, local entity groups, and chain connectivity at once, so it measures the structured encoder rather than one component.

\textbf{Language conditioning.}
Removing the action-type and modifier conditions reduces held-out SR from 80.0\% to 8.3\%, showing that 
explicit instruction conditioning is critical for robust semantic recombination.
Conditioning on the full subtask instruction without action/modifier decomposition still achieves 77.2\% held-out SR, 
indicating that the strong generalization of our method does not depend solely on explicit semantic factorization.

\textbf{Coordinate and output representations.}
Using absolute point coordinates behaves like removing RPC: seen-modifier performance is preserved while held-out performance drops sharply, indicating that the relative frame matters most when an object must be recombined with an unseen relation.
Point trajectories and Cartesian delta actions perform comparably as the output format; we keep point trajectories because they stay in the same TCP-relative geometry as the input and, unlike action labels, can be extracted from human videos, which supports transfer from human demonstrations~\cite{haldar2025point}.

\section{Conclusion}
We introduced CoMani, a benchmark that isolates modifier, action-type, and sequential composition under matched initial scenes, 
and GraphPoint, a framework that turns a semantic entity graph into future gripper point trajectories and resolves them into executable poses.
Experiments on CoMani show that GraphPoint responds to changes in the instructed relation and action under matched visual scenes, 
and that one atomic policy can be reused to complete multi-step instructions that were never demonstrated as a whole, 
confirming its instruction-dependent generalization at both the atomic and sequential levels.
The ablations further support the roles of entity separation, chain-mediated interaction, 
role-aware point collapse, language conditioning, and relative coordinates in this behavior.
The main limitations lie in the front-end perception modules, whose parsing, localization, 
and segmentation errors can propagate into control, and in severe occlusion, which can disrupt the tracked entity points.
In the future, we plan to explore obstacle avoidance for objects outside the task-relevant set, 
together with transfer to other embodiments and from human videos.

\bibliographystyle{IEEEtranBST/IEEEtran}
\bibliography{IEEEtranBST/ref}

\end{document}